\documentclass[conference]{IEEEtran}
\IEEEoverridecommandlockouts
\usepackage{cite}
\usepackage{amsmath,amssymb,amsfonts}
\usepackage{graphicx}
\usepackage{textcomp}
\usepackage[cmyk]{xcolor}

\usepackage{subfigure}
\usepackage{caption}
\usepackage{booktabs}
\usepackage[linesnumbered,ruled]{algorithm2e}
\usepackage{mathtools}
\usepackage{latexsym}
\usepackage{array}
\usepackage{multirow}
\usepackage{longtable}
\usepackage{rotating}

\begin{document}
\title{FedCME: Client Matching and Classifier Exchanging to Handle Data Heterogeneity in Federated Learning}



\author{
	\IEEEauthorblockN{
		Jun Nie\IEEEauthorrefmark{4}, 
		Danyang Xiao\IEEEauthorrefmark{4}, 
		Lei Yang\IEEEauthorrefmark{2}, 
		Weigang Wu$^{*,}$\IEEEauthorrefmark{4}}
	\IEEEauthorblockA{\IEEEauthorrefmark{4}School of Computer Science and Engineering, Sun Yat-sen University\\ \{niej7, xiaody\}@mail2.sysu.edu.cn, wuweig@mail.sysu.edu.cn}
	\IEEEauthorblockA{\IEEEauthorrefmark{2}School of Computer Science and Engineering, South China University of Technology\\ sely@scut.edu.cn}
} 
\maketitle
\begin{abstract}
Data heterogeneity across clients is one of the key challenges in Federated Learning (FL), which may slow down the global model convergence and even weaken global model performance.
Most existing approaches tackle the heterogeneity by constraining local model updates through reference to global information provided by the server.
This can alleviate the performance degradation on the aggregated global model. Different from existing methods, we focus the information exchange between clients, which could also enhance the effectiveness of local training and lead to generate a high-performance global model.
Concretely, we propose a novel FL framework named FedCME by client matching and classifier exchanging.
In FedCME, clients with large differences in data distribution will be matched in pairs, and then the corresponding pair of clients will exchange their classifiers at the stage of local training in an intermediate moment. Since the local data determines the local model training direction, our method can correct update direction of classifiers and effectively alleviate local update divergence.
Besides, we propose feature alignment to enhance the training of the feature extractor.
Experimental results demonstrate that FedCME performs better than FedAvg, FedProx, MOON and FedRS on popular federated learning benchmarks including FMNIST and CIFAR10, in the case where data are heterogeneous.
\end{abstract}

\begin{IEEEkeywords}
Federated learning, data heterogeneity, client matching, feature alignment
\end{IEEEkeywords}

\section{Introduction}
Federated Learning(FL)
\cite{DBLP:conf/mlsys/BonawitzEGHIIKK19,DBLP:journals/corr/abs-1912-04977,DBLP:conf/aistats/McMahanMRHA17}  
has emerged as a new paradigm of distributed machine learning, which enables multiple clients to collaboratively learn a powerful global model without transmitting local private data to the server.
It is now successfully used in some real-world scenarios, e.g., health care
\cite{ DBLP:conf/cvpr/Liu00DH21}, 
 smart city
\cite{DBLP:conf/globecom/QolomanyAA020, DBLP:journals/connection/ZhengZSWLL22} and recommended system
\cite{DBLP:journals/corr/abs-1811-03604, DBLP:journals/corr/abs-1911-11807}.

\par
Although federated learning has made great achievements in some scenarios, it still faces many challenges
\cite{DBLP:journals/spm/LiSTS20}, such as heterogeneity, communication cost and privacy protection. Among them, heterogeneity is divided into system heterogeneity and data heterogeneity. The former is when each client has a different amount of bandwidth and computational power, which can been partly resolved by native asynchronous scheme of federated learning
\cite{DBLP:journals/corr/abs-2303-08322,DBLP:conf/icc/NishioY19}.
In our work, we mainly focus on data heterogeneity, namely Non-IID problem, where clients have varying amounts of data coming from distinct distributions
\cite{ DBLP:conf/icml/HsiehPMG20}.
Because of data heterogeneity, the local training trajectory may diverge a lot from the global target due to the differences between the local and global data distribution
\cite{DBLP:journals/corr/abs-1806-00582}.
The more heterogeneous the local data set is, the slower the training convergence speed will be. What is more, it will be harder to attain a global model with good performance \cite{DBLP:journals/corr/abs-1909-06335}. This is because the local model is trained on its local data, which is achieved by  minimizing the local empirical loss. However, minimizing the local empirical loss is fundamentally inconsistent with minimizing the global empirical loss in heterogeneous federated learning
\cite{DBLP:conf/iclr/AcarZNMWS21, DBLP:conf/iclr/LiHYWZ20, DBLP:conf/icml/MalinovskiyKGCR20}.

\begin{figure}[t]
\centerline{\includegraphics[width=9cm]{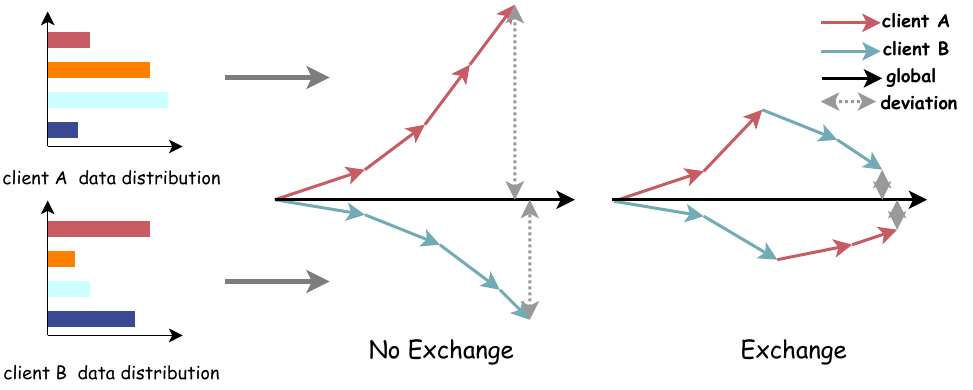}}
\caption{Model divergence in local training. This picture shows the model deviation due to data heterogeneity and shows the effect of exchanging and not exchanging.}
\label{deviation}
\end{figure}

\par
To address the data heterogeneity problem, quite a number of methods have been proposed. For instance,
FedProx \cite{DBLP:conf/mlsys/LiSZSTS20} 
puts forward to utilize a proximal term in the local training process to constrain the direction of local model updates, thereby reducing the gap between local and global optimization. MOON \cite{DBLP:conf/cvpr/LiHS21} uses model comparison to maximize the consistency between the representation learned by the current local model and the representation learned by the global model to correct local updates.
FedRS \cite{DBLP:conf/kdd/LiZ21} proposes
'Restricted Softmax' to limit the update of missing classes’ weights
during the local procedure. But in cases where there are no missing categories in the local data but large differences in the numbers between categories, the effect of FedRS will approximate that of FedAvg \cite{DBLP:conf/aistats/McMahanMRHA17}. Instead, we focus the information exchange between clients, which could
also enhance the effectiveness of local training and lead to get
a well-performed global model.

\par
Meanwhile, according to existing researches \cite{DBLP:conf/kdd/LiZ21},  the classifier has a greater impact on model performance than the feature extractor. In the work \cite{DBLP:conf/nips/LuoCHZLF21}, they find that the difference between the same model after training with different sets of heterogeneous data mainly lies in the classifier. This is also a significant reason why the training direction of the client model deviates from the training direction of the global model in the scenario of heterogeneous data. 

\par
Inspired by the observation above, we propose to handle data heterogeneity by classifier exchanging: the local model divergence is mitigated when the classifier is trained on two datasets with complementary distributions (as Figure \ref{deviation} shows) . However, how to find suitable client pairs is a key issue. In our method FedCME, the server will perform pairwise matching based on the latest evaluation vectors of clients (obtained by the client's self-evaluation, which roughly reflects the data distribution) at the beginning of each global iteration round. Afterwards halfway through the local training, the pairwise clients exchange classifiers with each other. The rest of the local training is conducted using the classifier from its counterpart.     

\par
Additionally, we propose feature alignment to assist local training. During local training, the features output by the feature extractor of the local model are aligned with their corresponding global features for each category to enhance the training of the feature extractor.

\par
Extensive experiments demonstrate that our method can have high training efficiency and reach better performance compared with some existing algorithms, FedAvg, FedProx, MOON, FedRS, using datasets including FMNIST, CIFAR10. Besides, we also conducted multiple ablation experiments to prove the rationality and effectiveness of our method.
\par
The rest of this paper is divided into five sections.
We present related works on addressing data heterogeneity in Section \uppercase\expandafter{\romannumeral2}. In Section \uppercase\expandafter{\romannumeral3}, FedAvg and data heterogeneity will be covered in detail. The methodology and experiments are presented in Section \uppercase\expandafter{\romannumeral4} and \uppercase\expandafter{\romannumeral5} respectively. Section \uppercase\expandafter{\romannumeral6} is the conclusion and outlines future work.

\section{Related Work}
The study in \cite{DBLP:conf/icml/KarimireddyKMRS20} demonstrates that data heterogeneity can slow down FL convergence speed. Furthermore, the performance of the final converged global model will be reduced as a result \cite{ DBLP:conf/icde/LiDCH22,DBLP:journals/corr/abs-1806-00582}.
Therefore, many methods have been proposed to solve this problem. They can be roughly divided into four categories:
    \par
    \textbf{Data Sharing.} These methods introduce public datasets or
    synthesized data to help construct a more balanced data distribution on the client or on the server. The conventional approach involves creating an adequate number of publicly shared datasets among clients \cite{DBLP:journals/corr/abs-1806-00582}.
    \par
    \textbf{Aggregation Scheme.}  These methods mainly improve the way in which the model is aggregated on the server side. For example, FedNova \cite{DBLP:conf/nips/WangLLJP20} 
    considers that clients with different computing power may need to perform varying numbers of local steps in local training.
    \par
    \textbf{Personalized Federated Learning.} Those methods aim to train personalized models for individual clients rather than a shared global model 
 \cite{DBLP:conf/nips/DinhTN20}, either by treating each
    client as a task in meta-learning  
 \cite{DBLP:conf/nips/0001MO20} or multi-task learning \cite{DBLP:journals/tnn/SattlerMS21}. 

    \par
    \textbf{Client Drift Mitigation.} Due to heterogeneous data, the optimization direction of the global model is inconsistent with that of the local model, resulting in what is called 'client drift' \cite{DBLP:conf/icml/KarimireddyKMRS20}. In order to mitigate it, a lot of targeted works has been produced. The first work among them is FedProx \cite{DBLP:conf/mlsys/LiSZSTS20}, which proposes a a proximal term to constrain the local model to deviate too much from the global model during training. Apart from this, MOON \cite{DBLP:conf/cvpr/LiHS21} aims to maximize the consistency between the representation learned by the local model and that learned by the global model through a contrastive loss. In addition, FedRS \cite{DBLP:conf/kdd/LiZ21}   takes into account more details regarding the local models. Specifically, it reveals that the top layers of neural networks are more task-specific, and good performance can also be achieved by only adjusting  the classifier of the local model based on local data distribution.
    \par
    Our work focuses on mitigating client drift. In our method, the model is divided into a feature extractor and a classifier during local training. For the feature extractor, we align local features with global features to improve its performance. For the classifier, we use evaluation vectors obtained through local model self-evaluation for doing clients matching and then exchange the classifier with their respective counterparts. 

\section{Preliminary}
In this section, we firstly introduce the most representative FL framework FedAvg, followed by an introduction to data heterogeneity and its categories.

\subsection{Federated Averaging Learning}
Federated Learning(FL) coordinates multiple clients with a central server to train a shared global model iteratively \cite{DBLP:journals/corr/abs-1912-04977}. The pioneer work is FedAvg \cite{DBLP:conf/aistats/McMahanMRHA17} and subsequent methods are based on it to make improvements. The framework of FedAvg consists of two parts: a server and clients. Let $ \mathcal{K} =\{1,2,3,...,K\}$ denotes the set of K clients, each of which has a local dataset $\mathcal{D}_{k \in \mathcal{K}}$. Each data sample $i$ in $\mathcal{D}_{k}$ can be represented by $\xi_{i}=\{\mathbf{x}_{i}, y_{i}\}$, where $\mathbf{x}_{i}$ is the $i$th data sample and $y_{i}$ is the label of the sample. For client k, it holds $D_{k} = |\mathcal{D}_{k}|$ data samples. For the entire federated learning system, there are total $D= {\textstyle \sum_{k=1}^{K}{D}_{k}}$ data samples. We use $l(w;\xi_{i})$ to denote the loss value given by the data sample $\xi_{i}$ on the model whose parameter is $w_k$. Then in FL, the objective is to minimize the total weighted loss:
\begin{equation}
    \min_{w} L(w) = { \sum^{K}_{k=1} \frac{D_{k}}{D}L_k(w)}.
\end{equation}
Where $L_k(w) = \frac{1}{D_k} {\textstyle \sum_{i \in \mathcal{D}_k}l(w;\xi_i)}$. In more detail, the process of FL training is composed with three parts.

\par \textbf{Step 1: Initialization and Select clients.}  The server will select a subset of $\mathcal{K}$,  $\mathcal{M}$. And client $k \in \mathcal{M}$ will participate in this round of training. Then the server transmits the model parameters to selected clients for their local training.

\par \textbf{Step 2: Local training.} In the t-th round of global training, the local training in selected clients usually employs stochastic gradient descent(SDG) \cite{DBLP:journals/ml/QianJY0Z15} method via using mini-batches samples from their local datasets,i.e.,
\begin{equation}
    w^{t+1}_k = w^{t}_k - \eta(\frac{1}{|\mathcal{B}_k|} \sum_{i \in \mathcal{B}_k} \bigtriangledown l(w^t_k;\xi_i)).
\end{equation}
Where $\eta$ is the learning rate, and $\mathcal{B}_k$ is the training data for once mini-batch SGD in client k. 

\par \textbf{Step 3: Aggregation.} After selected clients accomplish local training, they will send their model parameters to the server. Then, the server will aggregate clients to generate the new global model $w^{t+1}$ used for next round. As below:
\begin{equation}
    w^{t+1} = \frac{1}{D} \sum_{k \in \mathcal{M}} D_kw^{t+1}_k .  
\end{equation}

\par Then repeating 1-3 steps until the global loss function converges, or the test accuracy reaches the preset value.

\subsection{Data Heterogeneity}
Most of the existing researches \cite{DBLP:journals/corr/abs-1912-04977} divide data heterogeneity into five categories: feature distribution skew, label distribution skew, same label and different features, same features and different label, and quantity skew. Label distribution skew means $\mathcal{P}_k(y)$ may vary across clients in the situation of  same $\mathcal{P}(\mathbf{x}|y)$. And quantity skew denotes different clients can hold vastly different amounts of data.

\begin{figure*}[t]
\centerline{\includegraphics[width=16cm]{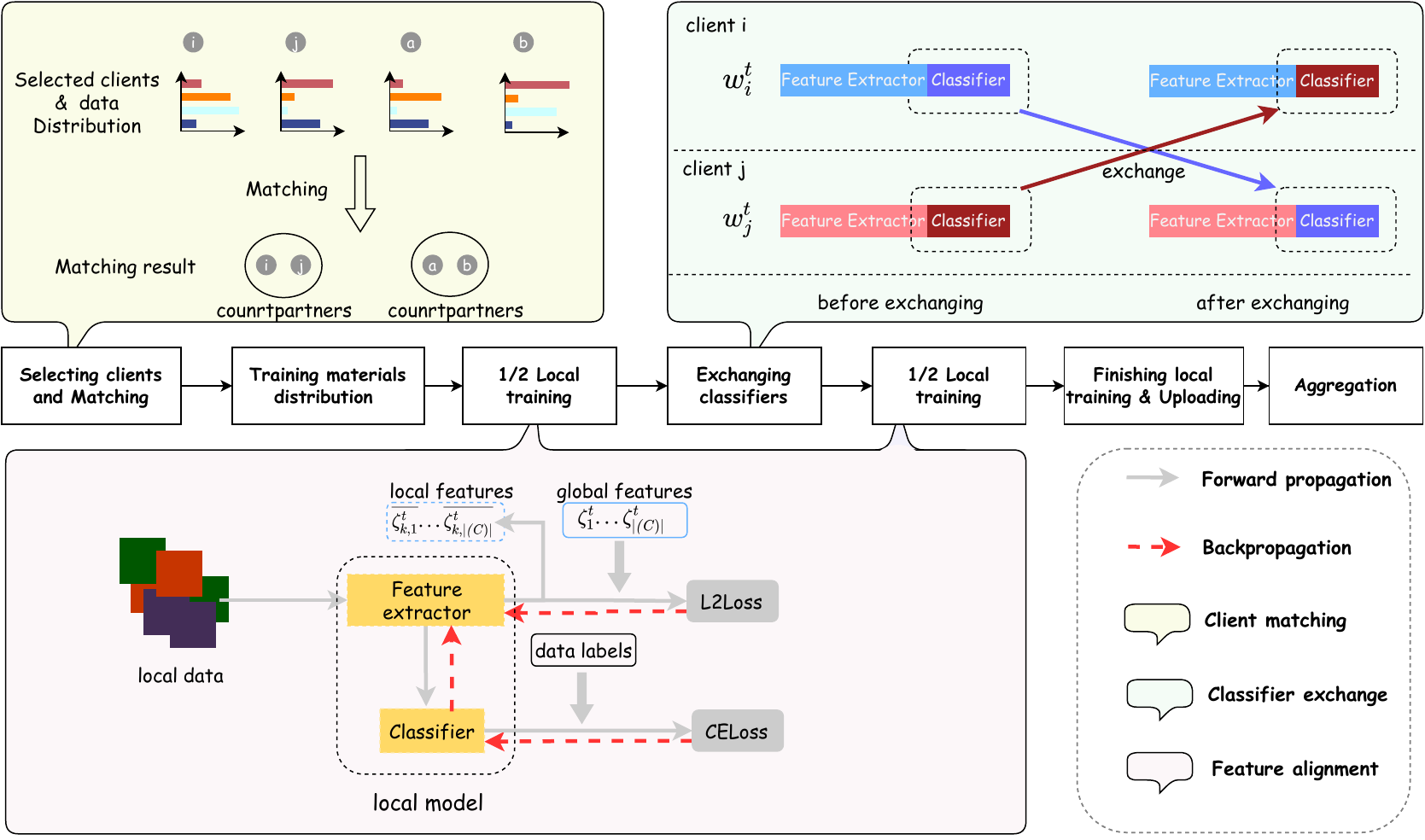}}
\caption{This figure illustrates the sequential steps involved in a single round of global iteration in FedCME. }
\label{FedCME-process}
\end{figure*}

\par In this paper, similar to previous works \cite{DBLP:conf/cvpr/LiHS21,DBLP:conf/mlsys/LiSZSTS20,DBLP:conf/kdd/LiZ21}, we focus on scenes involving a mixture of label distribution skew and quantity skew. In other words, a client may have a subset of data types and the number of its local data will differ from others'. 

\section{Methodology}
In this section, we introduce FedCME, which utilizes  classifier exchange and feature alignment to mitigate the slowdown in global model convergence and degradation in global model performance caused by data heterogeneity in FL.

\begin{table}
  \centering
  \caption{Main Notations and Definitions}
  \begin{tabular}{ll}
    \toprule
    Notation                      & Definition     \\
    \midrule
    $t$                         & the t-th global round \\
    $w^{t}$                     & the global model of the t-th global round \\
    $\theta$                    & the feature extractor of the training model \\
    $\varphi$                   & the classifier of the training model \\
    $\mathcal{C}$               & the categories of training samples \\
    $\eta$                      & learning rate \\
    $\mu$                       & L2 regularization factor \\
    $\mathcal{B}$               & mini-batch size \\
    $\mathcal{K}$               & all clients  \\
    $\mathcal{M}$               & the set of clients selected in  the t-th global round \\
    $\zeta^{t}$                 & the global features in the t-th gloabl round \\
    $\varepsilon^{t} $          & the evaluation vectors of clients in the t-th global round \\
    $\mathcal{D}_k$             & the local data of client $k$ \\
    $E$                         & the number of local training rounds \\
    $T$                         & the number of global training rounds \\
    \bottomrule
  \end{tabular}
  \label{symbols}
\end{table}

\par As with other methods \cite{DBLP:conf/cvpr/LiHS21, DBLP:conf/mlsys/LiSZSTS20, DBLP:conf/kdd/LiZ21, DBLP:conf/aistats/McMahanMRHA17}, the training goal of FedCME is to train an efficient global model through several rounds of global iterations.
The procedure in each global iteration for a round of global iteration is shown in Figure \ref{FedCME-process}.
At the beginning of each global round, FL server selects a subset clients of $\mathcal{K}$, denoted by $\mathcal{M}$. Client $k \in \mathcal{M}$ performs local training procedure for $E$  rounds. During local training, in addition to training its local model by using its local data, client $k$ also records the local data features extracted from the model feature extractor during the training process. Specially, client $k$ exchanges its classifier with its counterpart halfway through local training (client $k$ receives three things for this round: the global model, the information of its counterpart which is another client participating in this global round, and global features used to assist in training the local feature extractor of the model). And then it proceeds to complete the remaining half of the local training. After all selected clients finish local training, the FL server will receive the parameters of their local models for aggregation to obtain the global model of the next global round. It also receives local sample features which will be combined with global features from the current global round to generate new global features for the next round. Additionally, it also receives the evaluation vector obtained by self-evaluation using partial local data, which can be used to match with a counterpart when selected again. The server and clients repeat the above processes until the global model converges.  More details will be introduced below.

\subsection{Client Matching and Classifier Exchanging}
 Client matching is performed on the server side by using evaluation vectors at the beginning of each round of global iteration, and relevant details regarding the function \textbf{MakeMatching} can be found in the 8-th line of Algorithm \ref{FedCME}.  After the matching results are obtained, the client performs classifier exchanging on the client side at the middle moment of client local training in the current round of global iteration, and 
 the 22-th line of Algorithm \ref{ClientUpdate} can be referred to for the specific process.

\begin{algorithm}[!ht]
\caption{ServerExecute in FedCME}
\label{FedCME}
\LinesNumbered

\KwIn{a set of clients participating in FL, $\mathcal{K}=\{1,2,...,K\}$, a set of clients selected in each global round $\mathcal{M}$, the number of global epochs $T$, 
the global model $w^t$, global features $\zeta^t$ and the global evaluation  vectors $\varepsilon^{t}$ in the t-th global round}

\KwOut{the global model $w^{t+1}$, global features $\zeta^{t+1}$, the evaluation vector $\varepsilon^{t+1}$ for (t+1)-th global round }

\textbf{ServerExecute:}\\
	\For{each round t in T}{
		$\mathcal{M} \leftarrow random(\mathcal{K})$ \\
            dict $\nu^{t}$  $\leftarrow$ MakeMatching($\mathcal{M}$, $\varepsilon^{t}$) \\
            \For{each client $k$ in $\mathcal{M}$}{
            $w^t_{k}$, $\zeta^t_{k}$, $\varepsilon^t_{k}$   $\leftarrow$ ClientUpdate($w^t, \zeta^t, \nu^{t}[k]$ ) \\
            }
           $w^{t+1}$, $\zeta^{t+1}$, $\varepsilon^{t+1}$ $\leftarrow $Aggregation($\mathcal{M}$) \\
	}
\Return $w^{T}$

\textbf{MakeMatching}($\mathcal{M}$, $\varepsilon^{t}$): \\
initialize $\nu^{t}$, $\varepsilon^{*}$, $\varepsilon_{g}$ \\
\ForEach{client $k$ in $\mathcal{M}$}{
    $\varepsilon^{*}[k]$ $\leftarrow$ $\varepsilon^{t}[k]$ \\
    $\varepsilon_{g}$ $\leftarrow$ $\varepsilon_{g}$+$\varepsilon^{t}[k]/|\mathcal{M}|$ \\ 
}
sort vectors collection $\varepsilon^{*}$ by similarity $(\varepsilon^{*}[k],\varepsilon_{g})$ ascending \\
\While{$\varepsilon^{*}$ is not empty}{
    pop the first element from $\varepsilon^{*}$, denoted by $\varepsilon^{*}[k]$
    find the $j$, s.t. : \\
    $j = min (Similarity(\varepsilon^{*}[k], \varepsilon^{*}[j]), \  s.t. \ \varepsilon^{*}[j] \in \varepsilon^{*}$ \\
    $\nu^{t}[k], \nu^{t}[j] \leftarrow$ $j, k$ \\
    pop $\varepsilon^{*}[j]$ from $\varepsilon^{*}$ \\ 
}
\Return $\nu^{t}$

\textbf{Aggregation}($\mathcal{M}$): \\
$w^{t+1} \leftarrow  {\sum_{k \in \mathcal{M}}\frac{|\mathcal{D}_k|}{ |\mathcal{D}|}} w^{t}_k$ \\ 
\ForEach{client $k$ in  $\mathcal{M}$}{
    $\varepsilon^{t}[k]$ $\leftarrow$ $\varepsilon^{t}_{k}$ \\
    \ForEach{$c$ in $\mathcal{C}$}{
        \If{$\zeta^{t}_{k}[c]$ is empty}{
            $\zeta^{t}_{k}[c]$ $\leftarrow$ $\zeta^{t}[c]$
        }
    }
}
$\varepsilon^{t+1}, \zeta^{t+1}$ $\leftarrow$ $\varepsilon^{t},  \sum_{k \in \mathcal{M}} \zeta^{t}_{k}/ |\mathcal{M}| $ \\
\Return $w^{t+1}$, $\zeta^{t+1}$, $\varepsilon^{t+1}$
\end{algorithm}

\begin{algorithm}[ht]
    \caption{ClientExecute in FedCME}
    \label{ClientUpdate}
    \LinesNumbered

    \KwIn{the global model $w^{t}$, learning rate $\eta$, local training set $\mathcal{D}_k$ for client $k$, mini-batch data $\mathcal{B}$, the number of local epochs $E$, its counterpart client $j$, L2 regularization factor $\mu$}
    \KwOut{local model $w^{t}_k$, evaluation vector $\varepsilon^{t}_k$ and local features $\zeta^{t}_k$}

    \textbf{ClientUpdate}: \\
    initialize $\zeta^{t}_k $ \\
    \For{each round $e$ in $E$}{
        \If{ $e$ equal to $\left \lfloor E/2 \right \rfloor $ }{
         ExchangeClassifier($w^{t}_k, w^{t}_j$) \\   
        }
        \ForEach{ $\mathcal{B}$ in $\mathcal{D}^k$}{
        $\zeta \leftarrow$ $f(w^{t}_k(\theta;\mathcal{B}))$ \\
        $L2Loss$ $\leftarrow$ 0 \\
        \ForEach{$c$ in $\mathcal{C}$}{
        $\zeta_c \leftarrow \sum_{y=c} \zeta_{y},\quad \zeta_{y} \in \zeta $\\
        $\zeta^{t}_k[c]$ $\leftarrow $$\zeta^{t}_k[c]$ + $\zeta_c$ \\
        $L2Loss$ $\leftarrow $ L2Loss + $||(\zeta_c/ |\mathcal{B}_c|) -\zeta^{t}[c]||^{2}$  \\         
        }
        $w^{t}_k$ $\leftarrow w^{t}_k$ - $\eta * (\bigtriangledown f(w^{t}_k(\varphi );\zeta)+ \mu \bigtriangledown L2Loss$)  
        }
    }
    $\zeta^{t}_k[c] \leftarrow \zeta^{t}_k[c]/(E* |\mathcal{D}_{k,c}|)$, for $c\in \mathcal{C}$  \\
    $\mathcal{D}^{eval}_k$ $\leftarrow$ a subset  of $\mathcal{D}_k$ used for self-evaluation \\
    $\varepsilon^{t}_{k}[c]$ denotes the accuracy in $\mathcal{D}^{eval}_{k,c}$ for $ c \in \mathcal{C}$ to $w^{t}_{k}$ \\
    \Return $w^{t}_k, \zeta^{t}_k, \varepsilon^{t}_k$ \\
    \textbf{ExchangeClassifier}($w^{t}_{k}, w^{t}_{j}$): \\
    $\theta^t_k, \theta^t_j, \varphi^t_k, \varphi^t_j$ represent the features extractor and classifier of $w^{t}_{k}, w^{t}_{j}$ respectively \\
    then $w^{t}_{k}, w^{t}_{j}$ $\leftarrow w(\theta^t_k$, $\varphi^t_j$),  $w(\theta^t_j,\varphi^t_k)$  \\
\end{algorithm}

\textbf{Matching Mechanism.} Firstly, we introduce the evaluation vector $\varepsilon^{t}_{k}$ for client $k$ in the t-th global round, which may reflect the distribution of data on the client $k$. When client $k$ has finished its local training and got local model $w^{t}_{k}$, it randomly select a certain data set $\mathcal{D}^{eval}_k$ from $\mathcal{D}_{k}$ to evaluate $w^{t}_{k}$.
 For each $c$ in $\mathcal{C}_k$ (the set of categories in $\mathcal{D}^{eval}_k$), we can obtain the value corresponding to category c of the evaluation vector:
\begin{equation}\label{evalvct}
    \varepsilon^{t}_{k}[c] = \frac{\sum_{(x,y) \in {\mathcal{D}^{eval}_{k,c}}} g(w^{t}_{k};x;y)}{|\mathcal{D}^{eval}_{k,c}|} .
\end{equation}
Where $\mathcal{D}^{eval}_{k,c}$  is the data its label $c$  of $\mathcal{D}^{eval}_{k}$. 
 And $g(w^{t}_{k};x;y)$ is 1 if w correctly classified the sample $(x,y)$ and 0 otherwise. In other words knowing to (\ref{evalvct}), $\varepsilon^{t}_{k}[c]$ is the accuracy for $w^{t}_{k}$ on $\mathcal{D}^{eval}_{k,c}$, implying the model training offset direction. Notice that when $c$ in $\mathcal{C}$ but not in $\mathcal{C}_k$, $\varepsilon^{t}_{k}[c]$ is set to 0.
\par
At the beginning of each global round, the server will do client matching by using $\varepsilon^{t}$. And $\varepsilon^{t}$ records the last uploaded evaluation vectors for each client in $\mathcal{K}$. $\varepsilon^{*}$ is a subset of $\varepsilon^{t}$, which contains evaluation vectors of $\mathcal{M}$. Set $\varepsilon_{g}={\textstyle \sum_{\varepsilon \in \varepsilon^{*}} \varepsilon} / |\varepsilon^{*}|$. Then $\varepsilon^{*}$ is sorted by Cosine-Similarity \cite{DBLP:journals/corr/GunayAC14} between $\varepsilon_{g}$ and $\varepsilon^{*}_{i} (\varepsilon^{*}_{i} \in \varepsilon^{*})$ in ascending order. Then server repeats the following three steps until
$\varepsilon^{*}$ is empty: 
\begin{enumerate}
    \item Get the first evaluation vector $\varepsilon_{k}$ in $\varepsilon^{*}$.
    \item Find $\varepsilon_{j}$ (the evaluation vector of client $j$) from $\varepsilon^{*}$, satisfying  Cosine-Similarity($\varepsilon_k, \varepsilon_j$) is the minimum among $\varepsilon^{*}$.
    \item Then client $k$ and client $j$ become
     counterparts, and remove $\varepsilon_k$ and $\varepsilon_j$ from $\varepsilon^{*}$.
\end{enumerate}

\textbf{Classifiers Exchanging.}  Before the local training starts, the selected client $k$ receives the client matching result (its counterpart) in addition to the global model sent by the server. After $\left \lfloor E/2 \right \rfloor $ rounds local training on the client $k$, client $k$ will exchange the feature extractor of local model with its counterpart matched by FL server. Assuming client $k$ and client $j$ are counterparts,  $w(\theta^t_k,\varphi^t_k)$ and $w(\theta^t_j,\varphi^t_j)$ respectively represent their models when they are not exchanged, then the respective model parameters after exchanging are $w(\theta^t_k,\varphi^t_j)$ and $w(\theta^t_j,\varphi^t_k)$. 
After exchanging classifiers, they will continue with their respective local training until completing it.
Since the local model training direction is determined by the local data, and the exchanged classifier is trained by two approximately complementary data sets,    
the above operations can efficiently correct the direction of classifier training.

\subsection{Feature Alignment}
To assist with training, we propose a feature alignment method to enhance the feature extractor for local model training on the client side.  To be specific,  we align the local features obtained from the feature extractor of local model during local training with the global features of the corresponding category sent by the server. As a result, an additional loss function will be generated to perform  backpropagation for the feature extractor of local model. In the following we will describe in detail the local features and the global features, as well as the process of feature alignment.

\textbf{Global Features and Local Features.} For local features $\zeta^t_k$ of client $k$ in the t-th global round,
\begin{equation}
    \zeta^{t}_{k}[c] = \frac{\sum f(w^{t}_{k}(\theta);\mathcal{D}_{k,c})}{|\mathcal{D}_{k,c}|}, c \in \mathcal{C} .
\end{equation}
Where $\mathcal{D}_{k,c}$ denotes the data collection of category $c$ in the local data $\mathcal{D}_{k}$. The global features $\zeta^{t+1}[c]$ comes from aggregating all $\zeta^{t}_{k}$,  $k \in \mathcal{M}$. However, in data heterogeneity setting, client $k$ may own a subset of the set $\mathcal{C}$, represented by $\mathcal{C}_{k}$. In order to make global features more robust, we add a memory mechanism when we aggregate local features. That is, for client $k$ owning collection of sample categories $\mathcal{C}_{k}$, FL server will set $\zeta^{t}[c]$ = $\zeta^{t}_{k}[c]$ (where $c \in \mathcal{C}$  and $c \notin \mathcal{C}^{t}_{k})$. Then, for each $c$ in $\mathcal{C}$:
\begin{equation}
    \zeta^{t+1}[c] = \frac{\sum_{k \in \mathcal{M}}\zeta^{t}_{k}[c]}{| \mathcal{M}|} .
\end{equation}
    
This means that at the end of each global iteration, the server will update the global features through aggregation, in addition to getting the global model for the next global iteration through aggregation.

\textbf{Feature Extraction and Alignment.}  In FedAvg, the loss function used to update the local model at each step is defined as follows:
\begin{equation}
    CELoss = f(w^{t}_{k}(\theta,\varphi);\mathcal{B}_k) .
\end{equation}
\begin{equation}\label{backward}
    w^{t}_{k} \leftarrow w^{t}_{k} - \eta \bigtriangledown CELoss .
\end{equation}

Where $CELoss$  denotes Cross-Entropy loss \cite{DBLP:journals/bstj/Shannon48},  $w^{t}_{k}$ is the local model for client $k$,  $\mathcal{B}_k$ is the data for mini-batch SGD \cite{ DBLP:conf/icml/ZhengMWCYML17}. $\eta$ is the learning rate.
On the basis of the above, we add a loss, $L2Loss$, denoted the loss due to feature alignment. 
 As follows:
\begin{equation}
    L2Loss = \sum_{c \in \mathcal{C}}||\frac{\sum_{(x, y) \in \mathcal{B}_{k,c}} f(w^{t}_{k}(\theta);x) }{|\mathcal{B}_{k,c}|} - \zeta^{t}[c]||^{2}.
\end{equation}
 Then Eq.(\ref{backward}) turns into:
\begin{equation}
    w^{t}_{k} \leftarrow w^{t}_{k} - \eta (\bigtriangledown CELoss + \mu \bigtriangledown L2Loss).
\end{equation}
 Where $\theta$ is the feature extractor, and $\mathcal{B}_{k,c}$ is the data whose label is $c$ in $\mathcal{B}_{k}$. $\mu$ is L2 regularization factor.

\begin{table*}[t]
\centering
\caption{This table shows the test accuracy($\%$) of different FL  frameworks in different environments. $\alpha$ represents the coefficient of the dirichlet distribution and has two values of  0.1 and 0.5. There are four different scenarios with different proportions of client participation, such as K=50, M=10 means that in each round of global iteration, ten clients are selected from fifty clients to participate in this round of training. The parts with the highest and second highest accuracy in the table are bolded. Particularly, the highest accuracy is also underlined.}
\resizebox{\textwidth}{!}{%
\begin{tabular}{llllllllll}
\hline
                     & \multicolumn{4}{c}{$\alpha$=0.1}                                                                            &  & \multicolumn{4}{c}{$\alpha$=0.5}                                                                            \\ \cline{2-5} \cline{7-10} 
\multicolumn{1}{c}{} & \multicolumn{1}{c}{K=50, M=10} & \multicolumn{1}{c}{K=50, M=20} & \multicolumn{1}{c}{K=80, M=20} & \multicolumn{1}{c}{K=80, M=30} &  & \multicolumn{1}{c}{K=50, M=10} & \multicolumn{1}{c}{K=50, M=20} & \multicolumn{1}{c}{K=80, M=20} & \multicolumn{1}{c}{K=80, M=30} \\ \cline{2-10} 
                     & \multicolumn{9}{c}{FMNIST}                                                                                                                                                                                                       \\ \hline
FedAvg               & 85.84$\pm$0.2                & 86.05$\pm$0.29               & 85.64$\pm$0.21               & 86.05$\pm$0.11               &  & 88.82$\pm$0.02               & 88.81$\pm$0.09               & \textbf{89.26$\pm$0.01}      & 89.34$\pm$0.01               \\
FedProx(0.1)         & \textbf{\underline{86.14$\pm$0.28}}      & 85.94$\pm$0.38               & 85.71$\pm$0.19               & 85.84$\pm$0.1                &  & 88.69$\pm$0.03               & 88.87$\pm$0.06               & 89.16$\pm$0.01               & \textbf{89.35$\pm$0.02}      \\
FedProx(0.01)        & \textbf{85.99$\pm$0.22}      & \textbf{86.07$\pm$0.19}      & 85.85$\pm$0.22               & \textbf{86.21$\pm$0.13}      &  & 88.79$\pm$0.02               & 88.87$\pm$0.08               & 89.26$\pm$0.02               & 89.33$\pm$0.01               \\
MOON                 & 85.62$\pm$0.24               & 85.75$\pm$0.21               & \textbf{85.86$\pm$0.28}      & 86.06$\pm$0.11               &  & 88.63$\pm$0.12               & 88.91$\pm$0.05               & 89.18$\pm$0.01               & 89.25$\pm$0.01               \\
FedRS(0.1)           & 83.42$\pm$0.02               & 84.61$\pm$0.02               & 83.51$\pm$0.03               & 83.49$\pm$0.03               &  & \textbf{89.1$\pm$0.12}                & \textbf{89.33$\pm$0.06}      & 88.95$\pm$0.02               & 89.1$\pm$0.02                \\
FedRS(0.5)           & 84.43$\pm$0.01               & 85.38$\pm$0.04               & 84.51$\pm$0.02               & 84.59$\pm$0.05               &  & 88.89$\pm$0.11      & 89.25$\pm$0.01               & 89.17$\pm$0.01               & 89.1$\pm$0.01                \\
FedCME(0.1)          & 85.08$\pm$0.2                & 85.91$\pm$0.3                & 85.21$\pm$0.16               & 86.0$\pm$0.13                &  & 88.77$\pm$0.02               & 88.93$\pm$0.04               & 89.07$\pm$0.01               & 89.24$\pm$0.02               \\
FedCME(0.01)         & 85.59$\pm$0.12               & \textbf{\underline{86.26$\pm$0.15}}      & \textbf{\underline{86.01$\pm$0.25}}      & \textbf{\underline{86.4$\pm$0.16}}        &  & \textbf{\underline{89.52$\pm$0.03}}      & \textbf{\underline{89.57$\pm$0.06}}      & \textbf{\underline{89.69$\pm$0.02}}      & \textbf{\underline{89.84$\pm$0.03}}     \\ \hline
\multicolumn{1}{c}{} & \multicolumn{9}{c}{CIFAR10}                                                                                                                                                                                                      \\ \hline
FedAvg               & 64.08$\pm$0.41               & 65.98$\pm$0.55               & 62.64$\pm$0.05               & 65.16$\pm$0.31               &  & 74.85$\pm$0.44               & 75.19$\pm$0.12               & \textbf{75.3$\pm$0.1}                 & 74.95$\pm$0.04               \\
FedProx(0.1)         & 63.56$\pm$0.02               & 66.12$\pm$0.51               & 62.83$\pm$0.22               & 64.5$\pm$0.06                &  & 74.85$\pm$0.32               & 75.55$\pm$0.5                & \textbf{\underline{75.37$\pm$0.1}}       & 74.84$\pm$0.01               \\
FedProx(0.01)        & 64.38$\pm$0.02               & 66.04$\pm$0.53               & 62.62$\pm$0.2                & 64.86$\pm$0.28               &  & 74.83$\pm$0.35               & 75.65$\pm$0.49               & 75.22$\pm$0.08      & 75.07$\pm$0.01               \\
MOON                 & 63.89$\pm$0.68               & 65.5$\pm$0.61                & 61.47$\pm$0.25               & 64.88$\pm$0.6                &  & \textbf{75.02$\pm$0.39}      & \textbf{76.09$\pm$0.14}      & 74.89$\pm$0.04               & 74.91$\pm$0.03               \\
FedRS(0.1)           & 62.87$\pm$0.65               & 64.47$\pm$0.05               & 57.49$\pm$0.66               & 59.31$\pm$0.16               &  & 74.78$\pm$0.07               & 75.19$\pm$0.12               & 74.48$\pm$0.18               & 74.87$\pm$0.31               \\
FedRS(0.5)           & 62.88$\pm$0.5                & 64.29$\pm$0.08               & 57.91$\pm$0.56               & 59.7$\pm$0.26                &  & 74.01$\pm$0.22               & 74.93$\pm$0.02               & 74.56$\pm$0.11               & 74.84$\pm$0.22               \\
FedCME(0.1)          & \textbf{64.82$\pm$0.8}       & \textbf{66.87$\pm$0.18}      & \textbf{\underline{64.92$\pm$0.11}}      & \textbf{65.18$\pm$0.39}      &  & 74.38$\pm$0.48               & 75.86$\pm$0.52               & 75.19$\pm$0.08               & \textbf{\underline{75.23$\pm$0.02}}      \\
FedCME(0.01)         & \textbf{\underline{65.08$\pm$0.46}}      & \textbf{\underline{67.68$\pm$0.31}}      & \textbf{64.64$\pm$0.22}      & \textbf{\underline{67.64$\pm$0.25}}      &  & \textbf{\underline{75.19$\pm$0.63}}      & \textbf{\underline{76.37$\pm$0.7}}       & 75.1$\pm$0.12                & \textbf{75.19$\pm$0.06}      \\ \hline
\end{tabular}%
}
\label{acc}
\end{table*}

\begin{table*}[t]
\centering
\caption{The table shows that the global model test accuracy($\%$) of various FL frameworks at four global training time points on CIFAR10 with dirichlet distribution coefficient $\alpha$=0.1. T represents the total number of global rounds. The highest accuracy for each group is bolded, and the green digits below it indicates how much better than the highest accuracy among other methods without FedCME.}
\resizebox{\textwidth}{!}{%
\begin{tabular}{llllllllll}
\hline
              & \multicolumn{9}{c}{CIFAR10, $\alpha$=0.1}                                                                                                                                                                                                                                                                                                                                                                                                                                                                                                                                                                                                                                                                                                                            \\ \cline{2-10} 
              & \multicolumn{4}{c}{K=50, M=20}                                                                                                                                                                                                                                                                                                                                                &  & \multicolumn{4}{c}{K=80, M=30}                                                                                                                                                                                                                                                                                                                                                 \\ \cline{2-5} \cline{7-10} 
              & \multicolumn{1}{c}{1/5(T)}                                                                & \multicolumn{1}{c}{2/5(T)}                                                               & \multicolumn{1}{c}{3/5(T)}                                                                & \multicolumn{1}{c}{4/5(T)}                                                                &  & \multicolumn{1}{c}{1/5(T)}                                                                & \multicolumn{1}{c}{2/5(T)}                                                                & \multicolumn{1}{c}{3/5(T)}                                                                & \multicolumn{1}{c}{4/5(T)}                                                                \\ \hline
FedAvg        & 48.78$\pm$0.2                                                                                & 60.49$\pm$0.29                                                                              & 64.43$\pm$0.02                                                                               & 66.93$\pm$0.19                                                                               &  & 50.47$\pm$1.8                                                                                & 56.61$\pm$0.71                                                                               & 60.91$\pm$0.03                                                                               & 63.76$\pm$0.45                                                                               \\
FedProx(0.1)  & 48.87$\pm$0.15                                                                               & 59.72$\pm$0.23                                                                              & 64.62$\pm$0.25                                                                               & 66.66$\pm$0.04                                                                               &  & 49.53$\pm$1.33                                                                               & 56.43$\pm$0.73                                                                               & 60.51$\pm$0.04                                                                               & 63.94$\pm$0.61                                                                               \\
FedProx(0.01) & 48.69$\pm$0.19                                                                               & 60.42$\pm$0.3                                                                               & 64.74$\pm$0.16                                                                               & 66.65$\pm$0.03                                                                               &  & 50.3$\pm$1.36                                                                                & 56.93$\pm$0.79                                                                               & 60.85$\pm$0.05                                                                               & 63.6$\pm$0.43                                                                                \\
MOON          & 48.46$\pm$0.19                                                                               & 60.23$\pm$0.2                                                                               & 63.97$\pm$0.05                                                                               & 66.61$\pm$0.25                                                                               &  & 49.74$\pm$1.11                                                                               & 56.32$\pm$0.95                                                                               & 60.3$\pm$0.18                                                                                & 62.74$\pm$0.48                                                                               \\
FedRS(0.1)    & 51.72$\pm$0.46                                                                               & 60.33$\pm$0.04                                                                              & 63.9$\pm$0.2                                                                                 & 63.78$\pm$0.12                                                                               &  & 43.62$\pm$0.5                                                                                & 55.82$\pm$0.31                                                                               & 57.27$\pm$0.25                                                                               & 60.44$\pm$0.28                                                                               \\
FedRS(0.5)    & 52.4$\pm$0.31                                                                                & 60.46$\pm$0.06                                                                              & 64.05$\pm$0.28                                                                               & 63.84$\pm$0.02                                                                               &  & 44.05$\pm$0.52                                                                               & 56.33$\pm$0.36                                                                               & 57.91$\pm$0.24                                                                               & 60.02$\pm$0.21                                                                               \\
FedCME(0.1)   & 48.87$\pm$0.3                                                                                & 60.08$\pm$0.45                                                                              & 64.24$\pm$0.06                                                                               & 67.16$\pm$0.53                                                                               &  & 50.25$\pm$1.07                                                                               & 57.17$\pm$0.83                                                                               & 60.96$\pm$0.39                                                                               & 63.57$\pm$0.5                                                                                \\
FedCME(0.01)  & \multicolumn{1}{c}{\textbf{\begin{tabular}[c]{@{}c@{}}53.97$\pm$0.27 \\ \textcolor{green}{(1.57)$\uparrow$}\end{tabular}}} & \multicolumn{1}{c}{\textbf{\begin{tabular}[c]{@{}c@{}}63.2$\pm$0.05\\ \textcolor{green}{(2.71)$\uparrow$}\end{tabular}}} & \multicolumn{1}{c}{\textbf{\begin{tabular}[c]{@{}c@{}}65.99$\pm$0.02\\ \textcolor{green}{(1.25)$\uparrow$}\end{tabular}}} & \multicolumn{1}{c}{\textbf{\begin{tabular}[c]{@{}c@{}}67.59$\pm$0.63\\ \textcolor{green}{(0.66)$\uparrow$}\end{tabular}}} &  & \multicolumn{1}{c}{\textbf{\begin{tabular}[c]{@{}c@{}}55.23$\pm$1.27\\ \textcolor{green}{(4.76)$\uparrow$}\end{tabular}}} & \multicolumn{1}{c}{\textbf{\begin{tabular}[c]{@{}c@{}}61.83$\pm$0.69\\ \textcolor{green}{(4.9)$\uparrow$}\end{tabular}}} & \multicolumn{1}{c}{\textbf{\begin{tabular}[c]{@{}c@{}}66.03$\pm$0.64\\ \textcolor{green}{(5.12)$\uparrow$}\end{tabular}}} & \multicolumn{1}{c}{\textbf{\begin{tabular}[c]{@{}c@{}}66.22$\pm$0.32\\ \textcolor{green}{(2.28)$\uparrow$}\end{tabular}}} \\ \hline
\end{tabular}%
}
\label{train speed}
\end{table*}

\section{Experiments}
We conduct extensive experiments to verify the effectiveness of the proposed method and compare it with several classic and advanced methods in various datasets and settings (see the Appendix for more details).  Ablation studies are also conducted to verify the effectiveness of each component in FedCME and other related issues.

\subsection{Experimental results}


We conduct extensive experiments to demonstrate the superiority of FedCME in terms of model performance and training efficiency compared to other methods. Furthermore, experiments demonstrate the robustness and superiority of FedCME across different levels of participation and data heterogeneity.

    \textbf{Better Performance.} Table \ref{acc} reports the test accuracy of all compared algorithms on FMNIST and CIFAR10 datasets with various settings. We compare our method with other FL frameworks including: FedAvg \cite{DBLP:conf/aistats/McMahanMRHA17}, FedProx \cite{DBLP:conf/mlsys/LiSZSTS20}, MOON \cite{DBLP:conf/cvpr/LiHS21} and FedRS \cite{DBLP:conf/kdd/LiZ21}. For FedProx and FedRS, there are two settings for the coefficient of the local regularization term. On FMNIST, under eight different settings, FedCME with $\mu$=0.01 achieves the highest accuracy in seven. Especially,  FedCME with $\mu=0.1$ and $\mu=0.01$ achieves good results in all four settings on CIFAR10 with $\alpha =0.1$, occupying the highest accuracy and the second highest accuracy respectively. We can find FedCME with $\mu=0.01$ achieves the highest accuracy in most cases. What is more, FedCME performs better on CIFAR10 than on FMNIST. This shows that our method will improve more obviously than other methods when the training data has more extractable features. Particularly, the more heterogeneous the data is and the more our method's performance improves. Therefore, FedCME can effectively alleviate data heterogeneity compared with other frameworks.

    \textbf{Better Training  Efficiency.} Table \ref{train speed} compares the training speed of FedCME and other mentioned methods on CIFAR10 with $\alpha=0.1$. In both cases of client selection, FedCME with $\mu=0.01$ is always the best one to speed up FL training compared with other methods at the same time point. Specially, FedCME could achieve a target accuracy using fewer communication rounds than FedAvg, FedProx, MOON and FedRS. For instance, when K=80 and M=30, the test accuracy of FedCME with $\mu=0.01$ at $2/5(T)$ is about 61.83$\%$. At the same time point, FedProx(0.01) has the highest test accuracy among other methods without FedCME, which is about 56.93$\%$. FedCME with $\mu=0.01$ has a 4.9$\%$ higher test accuracy than FedProx(0.01). We attribute it to the fact that our method can effectively correct the client training direction in the data heterogeneous. Thence, our method has better training efficiency under same settings.

\subsection{Ablation Study}
In this section, we verify the effectiveness of each component in FedCME and explain why the matching process needs to use One-to-One matching instead of Many-to-One matching. Additionally, we organize two comparative experiments of exchanging the whole model and only exchanging the feature extractor. The following experimental results are attained  under two settings: (K=50, M=20) and (K=80, M=30) on CIFAR10 with $\alpha=0.1$ and $\mu=0.01$.

    \begin{figure}[htbp]
    	\centering
    	\begin{minipage}{0.49\linewidth}
    		\includegraphics[width=1\linewidth]{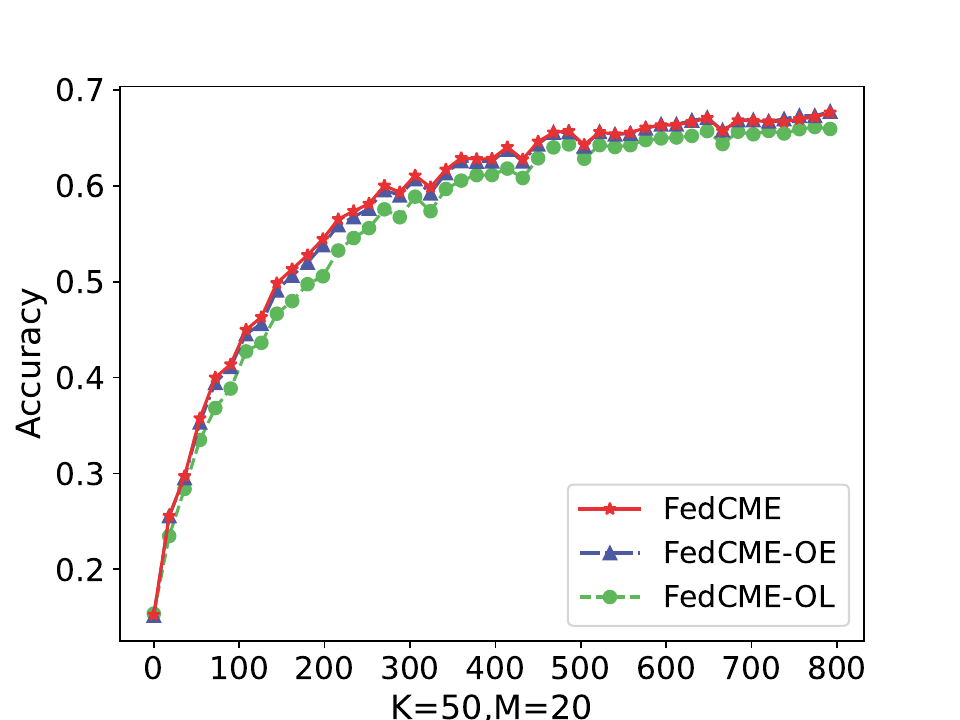}
    	\end{minipage}
    	\begin{minipage}{0.49\linewidth}
    		\includegraphics[width=1\linewidth]{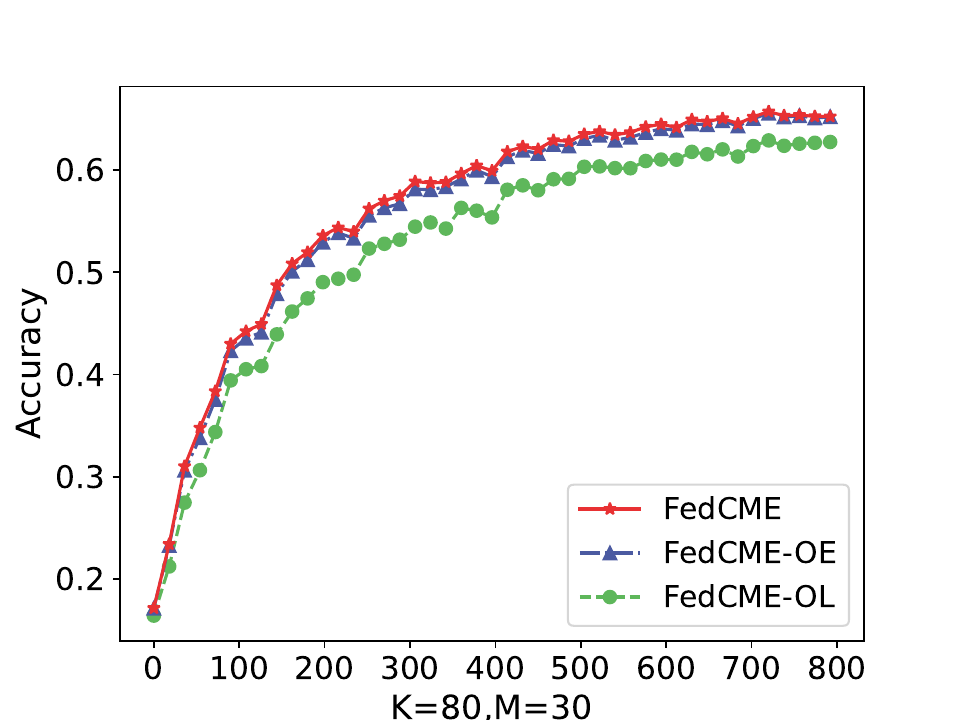}
    	\end{minipage}
        \caption{The two figures show the contribution of L2Loss and Exchanging to the FedCME in two different cases. The one on the left is K=50, M=20 and the one on the right is K=80, M=30. The tag FedCME-OL indicates that there is only \textit{L2Loss} in FedCME without \textit{Exchanging}. And FedCME-OE denotes FedCME with only \textit{Exchanging}.}
        \label{necessity of parts}
    \end{figure}

\begin{table}[htbp]
\centering
\caption{This table compares the test accuracy(\%) of FedCME with only classifier exchanging (FedCME-OE) and FedCME with only L2Loss (FedCME-OL) to other methods.}
\label{FedOL-FedOE}
\scalebox{1.1}{%
\begin{tabular}{lll}
\hline
                & \multicolumn{2}{c}{CIFAR10, $\alpha=0.1$}         \\ \cline{2-3} 
                & \multicolumn{1}{c}{K=50, M=20} & \multicolumn{1}{c}{K=80, M=30} \\ \hline
FedAvg          & $65.98\pm0.55 $                   & $65.16\pm0.31$                   \\ \hline
FedProx(0.1)    & $66.12\pm0.51 $                   & $64.5\pm0.06 $                   \\ \hline
FedProx(0.01)   & $66.04\pm0.53   $                 &$ 64.86\pm0.28 $                  \\ \hline
MOON            & $65.5\pm0.61  $                   & $64.88\pm0.6   $                 \\ \hline
FedRS(0.1)      & $64.47\pm0.05 $                   & $59.31\pm0.16  $                 \\ \hline
FedRS(0.5)      & $64.29\pm0.08$                    & $59.7\pm0.26   $                 \\ \hline
FedCME-OL(0.01) & $66.05\pm0.21   $                 & $65.21\pm0.13   $                \\ \hline
FedCME-OE(0.01) & $67.31\pm0.36   $                 & $67.24\pm0.1 $                   \\ \hline
FedCME(0.01)    & $67.68\pm0.31     $               & $67.64\pm0.25$                 \\ \hline
\end{tabular}%
}
\end{table}

    \begin{figure}[htbp]
	\centering
	\begin{minipage}{0.49\linewidth}
		\centering
		\includegraphics[width=1\linewidth]{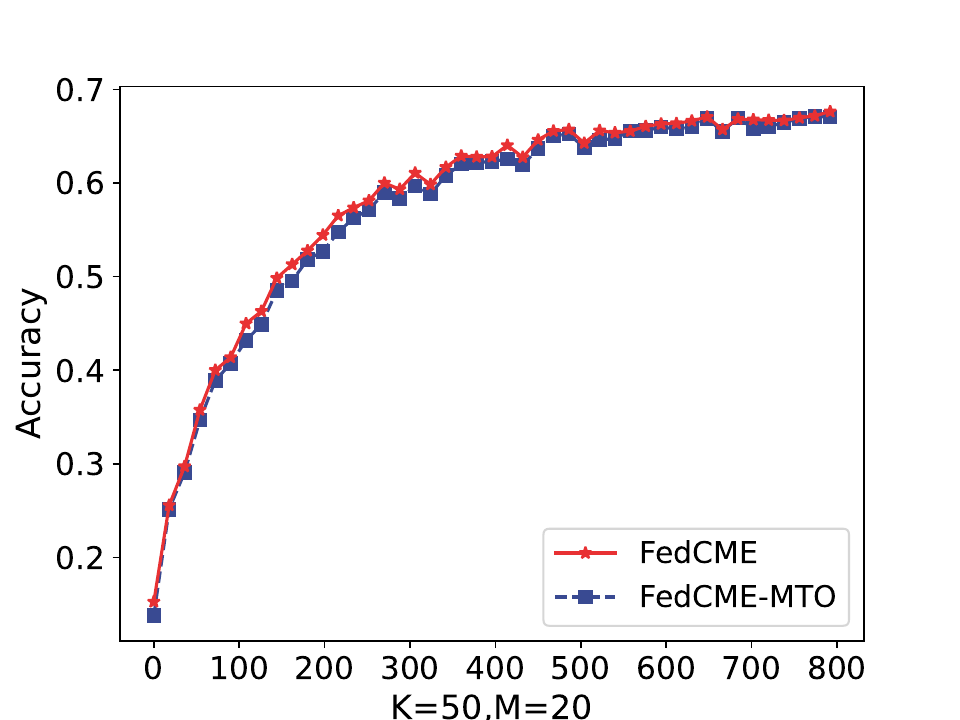}
	\end{minipage}
	\begin{minipage}{0.49\linewidth}
		\centering
		\includegraphics[width=1\linewidth]{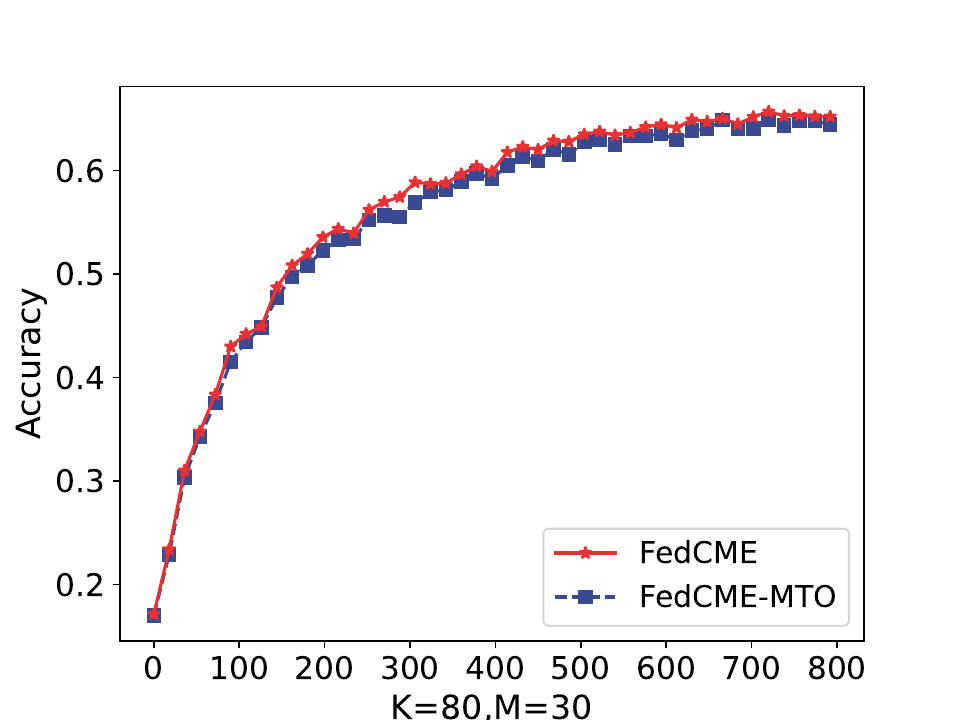}
	\end{minipage}
    \caption{These two figures show the training results in two matching modes. FedCME-MTO denotes the case of many-to-one matching.}
    \label{many to one matching}
    \end{figure}

    \textbf{Effectiveness of Each Component.} The parts where our method works consist of two: (1) the loss (denoted by \textit{L2Loss}) generated between the training sample extraction features and the global features in the local training and (2) update model direction correction produced by exchanging classifiers (denoted by \textit{Exchanging}). As we can see from Figure \ref{necessity of parts} and Table \ref{FedOL-FedOE}, \textbf{\textit{Exchanging} in FedCME plays a major role, and \textit{L2Loss} plays a supporting role}. Moreover, the influence of \textit{Exchanging} in FedCME runs through the entire training process. Unlike \textit{Exchanging}, \textit{L2Loss} is only effective in the first half of the global training process. This phenomenon is attributed to the gradual performance stabilization of the feature extractor \cite{DBLP:conf/nips/LuoCHZLF21} and the gradual convergence of global features.

    \begin{figure}[hbpt]
	\centering
	\begin{minipage}{0.49\linewidth}
        \centerline{\includegraphics[width=1\linewidth]{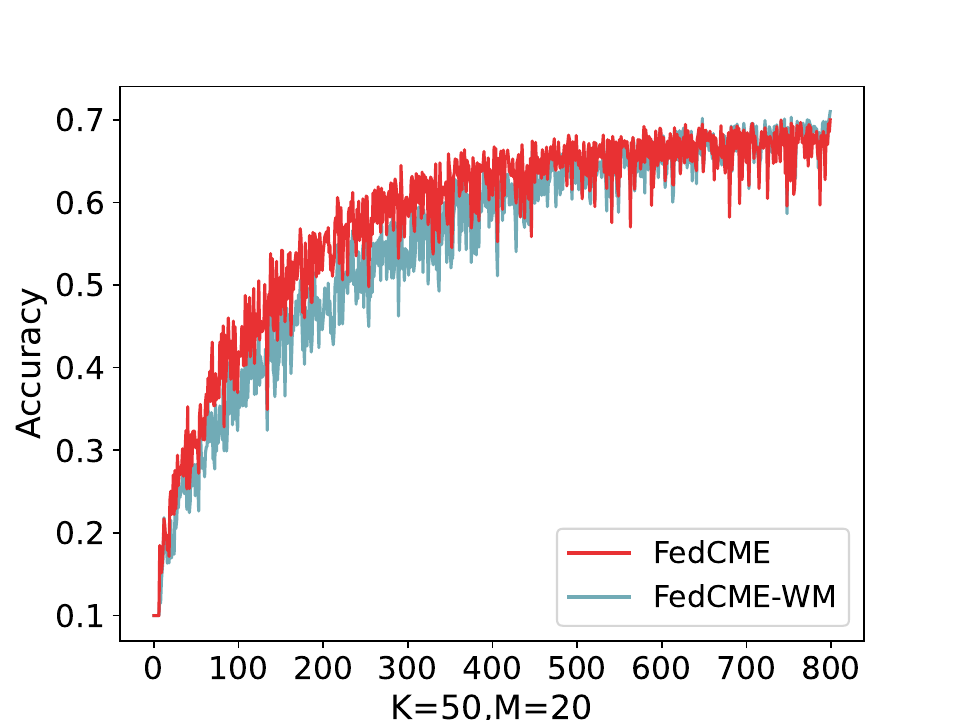}}
        \centerline{\includegraphics[width=1\linewidth]{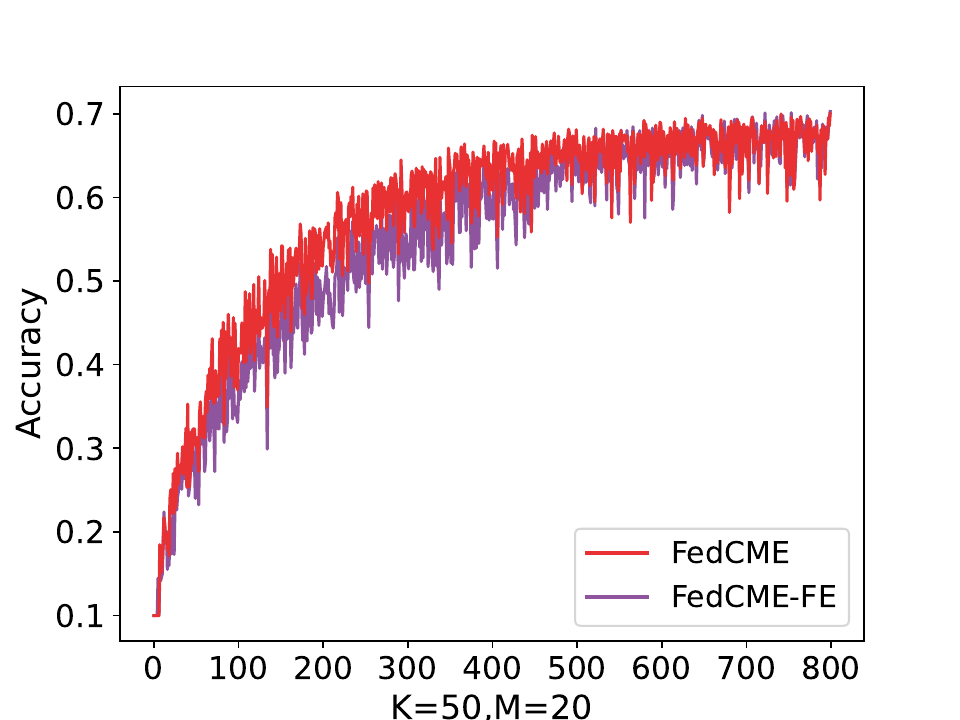}}
	\end{minipage}
	\begin{minipage}{0.49\linewidth}
        \centerline{\includegraphics[width=1\linewidth]{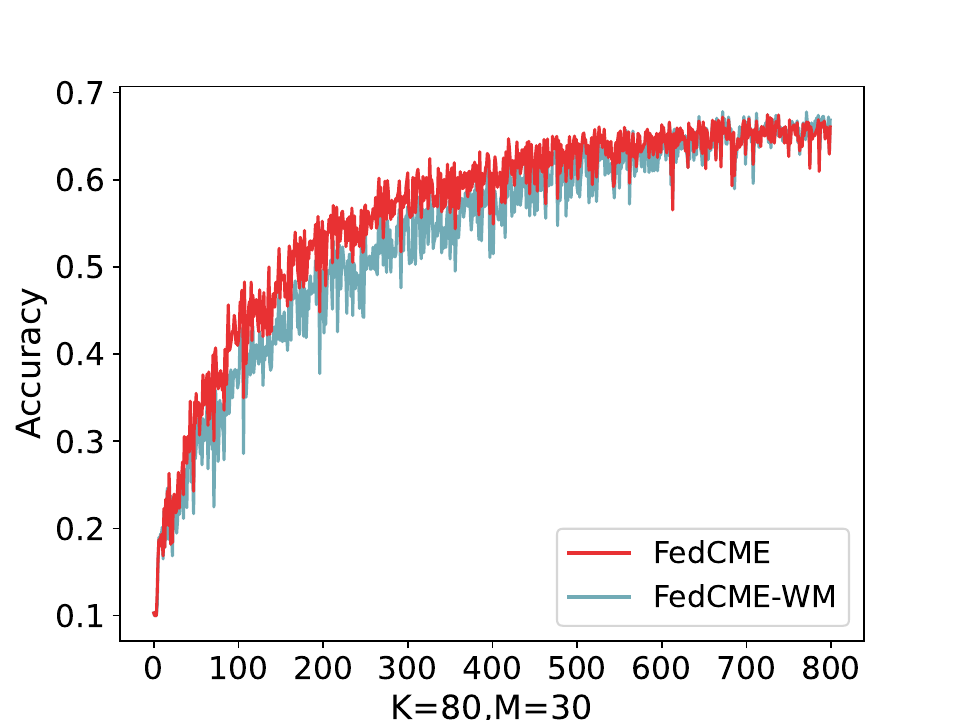}}
		\centerline{\includegraphics[width=1\linewidth]{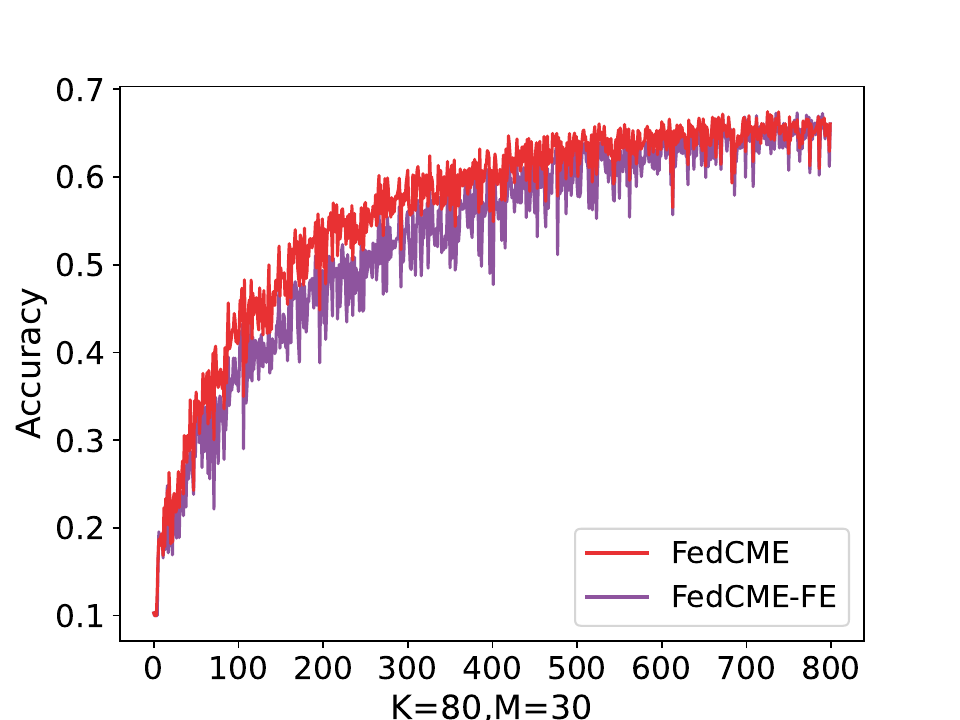}}
	\end{minipage}
    \caption{This four figures show four comparative experiments on exchanging the whole model (FedCME-WM) and only exchanging the feature extractor (FedCME-FE) compared with our method (FedCME).}
    \label{Why exchange classifier} 
    \end{figure}

    \textbf{How About Many-to-One Matching.} In our method, we take pairwise matching. Why not many-to-one matching?   To investigate this question, we conduct two comparison experiments using many-to-one matching between clients, in which clients can be matched repeatedly. During the matching process, multiple classifiers from different client models may be replaced by the same classifier that is deemed most suitable. The result is shown in Figure \ref{many to one matching}: many-to-one matching results are worse than pairwise matching results. In these experiments, we find that some clients are not matched during many-to-one matching, and the classifiers of these clients are not used. This causes the framework to lose some useful information.

\textbf{How About Exchanging The Whole Model Or The Feature Extractor.} As we can see in Figure \ref{Why exchange classifier}, when we choose to exchange the whole model parameters during exchanging, not only is it not better, but it also slows down the training of the model than exchanging the classifier. Therefore, this means that exchanging the whole model parameters is not as good as exchanging the classifier performance. Likewise, when only exchanging the feature extractor, the training is not as efficient as when exchanging the classifier. More notably, most models nowadays consist of one or two linear layers in the classifier, while the number of layers in the feature extractor is getting deeper and deeper in order to improve the model's effectiveness. This also indicates that there will be unnecessary communication overhead if the feature extractor needs to be exchanged.

\section{Conclusion and Future Work}
In this work we propose FedCME, a FL framework that enables FL to attain a more efficient global model in the case of heterogeneous data among clients.
In FedCME, we propose a  matching mechanism for exchanging classifiers to effectively mitigate the effects of data heterogeneity in the training process and propose feature alignment to assist the training process.
Extensive experiments demonstrate that FedCME provides better performance for mitigating client drift caused by data heterogeneity. Furthermore, in different degrees of data heterogeneity and different client selection strategies, FedCME also has a good performance.

In our current work, the model is statically divided into a feature extractor and a classifier, and the exchange time is also determined statically. However, there may be more efficient dynamic methods for different scenarios and models. In future work, we will explore this direction.

\bibliographystyle{ieeetr}
\bibliography{MSN-FedCME}

\begin{thebibliography}{10}

\bibitem{DBLP:conf/mlsys/BonawitzEGHIIKK19}
K.~A. Bonawitz and H.~E. et~al., ``Towards federated learning at scale: System
  design,'' in {\em Proceedings of Machine Learning and Systems, MLSys}, 2019.

\bibitem{DBLP:journals/corr/abs-1912-04977}
P.~Kairouz and H.~B.~M. et~al., ``Advances and open problems in federated
  learning,'' {\em Arxiv}, p.~abs/1912.04977, 2019.

\bibitem{DBLP:conf/aistats/McMahanMRHA17}
B.~McMahan and E.~M. et~al., ``Communication-efficient learning of deep
  networks from decentralized data,'' in {\em International Conference on
  Artificial Intelligence and Statistics, AISTATS}, p.~54, 2017.

\bibitem{DBLP:conf/cvpr/Liu00DH21}
Q.~Liu and C.~C. et~al., ``Feddg: Federated domain generalization on medical
  image segmentation via episodic learning in continuous frequency space,'' in
  {\em Computer Vision and Pattern Recognition, CVPR}, 2021.

\bibitem{DBLP:conf/globecom/QolomanyAA020}
B.~Qolomany and K.~A. et~al., ``Particle swarm optimized federated learning for
  industrial iot and smart city services,'' in {\em Global Communications
  Conference, GLOBECOM}, 2020.

\bibitem{DBLP:journals/connection/ZhengZSWLL22}
Z.~Zheng and Y.~Z. et~al., ``Applications of federated learning in smart
  cities: recent advances, taxonomy, and open challenges,'' {\em Connect.
  Sci.}, vol.~34, no.~1, pp.~1--28, 2022.

\bibitem{DBLP:journals/corr/abs-1811-03604}
A.~Hard and K.~R. et~al., ``Federated learning for mobile keyboard
  prediction,'' {\em Arxiv}, p.~abs/1811.03604, 2018.

\bibitem{DBLP:journals/corr/abs-1911-11807}
F.~Hartmann and S.~S. et~al., ``Federated learning for ranking browser history
  suggestions,'' {\em Arxiv}, p.~abs/1911.11807, 2019.

\bibitem{DBLP:journals/spm/LiSTS20}
T.~Li and A.~K.~S. et~al., ``Federated learning: Challenges, methods, and
  future directions,'' {\em {IEEE} Signal Process. Mag.}, vol.~37, no.~3,
  pp.~50--60, 2020.

\bibitem{DBLP:journals/corr/abs-2303-08322}
B.~Luo and X.~O. et~al., ``Optimization design for federated learning in
  heterogeneous 6g networks,'' {\em Arxiv}, p.~abs/2303.08322, 2023.

\bibitem{DBLP:conf/icc/NishioY19}
T.~Nishio and R.~Yonetani, ``Client selection for federated learning with
  heterogeneous resources in mobile edge,'' in {\em International Conference on
  Communications, ICC}, pp.~1--7, 2019.

\bibitem{DBLP:conf/icml/HsiehPMG20}
K.~Hsieh and A.~P. et~al., ``The non-iid data quagmire of decentralized machine
  learning,'' in {\em International Conference on Machine Learning, ICML},
  p.~119, 2020.

\bibitem{DBLP:journals/corr/abs-1806-00582}
Y.~Zhao and M.~L. et~al., ``Federated learning with non-iid data,'' {\em
  Arxiv}, p.~abs/1806.00582, 2018.

\bibitem{DBLP:journals/corr/abs-1909-06335}
T.~H. Hsu and H.~Q. et~al., ``Measuring the effects of non-identical data
  distribution for federated visual classification,'' {\em Arxiv},
  p.~abs/1909.06335, 2019.

\bibitem{DBLP:conf/iclr/AcarZNMWS21}
D.~A.~E. Acar and Y.~Z. et~al., ``Federated learning based on dynamic
  regularization,'' in {\em International Conference on Learning
  Representations, ICLR}, 2021.

\bibitem{DBLP:conf/iclr/LiHYWZ20}
X.~Li and K.~H. et~al., ``On the convergence of fedavg on non-iid data,'' in
  {\em International Conference on Learning Representations, ICLR}, 2020.

\bibitem{DBLP:conf/icml/MalinovskiyKGCR20}
G.~Malinovskiy and D.~K. et~al., ``From local {SGD} to local fixed-point
  methods for federated learning,'' in {\em International Conference on Machine
  Learning, ICML}, p.~119, 2020.

\bibitem{DBLP:conf/mlsys/LiSZSTS20}
T.~Li and A.~K.~S. et~al., ``Federated optimization in heterogeneous
  networks,'' in {\em Proceedings of Machine Learning and Systems, MLSys},
  2020.

\bibitem{DBLP:conf/cvpr/LiHS21}
Q.~Li and B.~H. et~al., ``Model-contrastive federated learning,'' in {\em
  Conference on Computer Vision and Pattern Recognition, CVPR}, 2021.

\bibitem{DBLP:conf/kdd/LiZ21}
X.~Li and D.~Zhan, ``Fedrs: Federated learning with restricted softmax for
  label distribution non-iid data,'' in {\em ACM Knowledge Discovery and Data
  Mining, SIGKDD}, 2021.

\bibitem{DBLP:conf/nips/LuoCHZLF21}
M.~Luo and F.~C. et~al., ``No fear of heterogeneity: Classifier calibration for
  federated learning with non-iid data,'' in {\em Neural Information Processing
  Systems, NeurIPS}, 2021.

\bibitem{DBLP:conf/icml/KarimireddyKMRS20}
S.~P. Karimireddy, S.~Kale, and M.~M. et~al., ``{SCAFFOLD:} stochastic
  controlled averaging for federated learning,'' in {\em International
  Conference on Machine Learning, ICML}, p.~119, 2020.

\bibitem{DBLP:conf/icde/LiDCH22}
Q.~Li and Y.~D. et~al., ``Federated learning on non-iid data silos: An
  experimental study,'' in {\em International Conference on Data Engineering,
  ICDE}, 2022.

\bibitem{DBLP:conf/nips/WangLLJP20}
J.~Wang and Q.~L. et~al., ``Tackling the objective inconsistency problem in
  heterogeneous federated optimization,'' in {\em Neural Information Processing
  Systems, NeurIPS}, 2020.

\bibitem{DBLP:conf/nips/DinhTN20}
C.~T. Dinh and N.~H.~T. et~al., ``Personalized federated learning with moreau
  envelopes,'' in {\em Neural Information Processing Systems, NeurIPS}, 2020.

\bibitem{DBLP:conf/nips/0001MO20}
A.~Fallah and A.~M. et~al., ``Personalized federated learning with theoretical
  guarantees: {A} model-agnostic meta-learning approach,'' in {\em Neural
  Information Processing Systems, NeurIPS}, 2020.

\bibitem{DBLP:journals/tnn/SattlerMS21}
F.~Sattler, K.~M{\"{u}}ller, and W.~Samek, ``Clustered federated learning:
  Model-agnostic distributed multitask optimization under privacy
  constraints,'' {\em {IEEE} Trans. Neural Networks Learn. Syst.}, vol.~32,
  no.~8, pp.~3710--3722, 2021.

\bibitem{DBLP:journals/ml/QianJY0Z15}
Q.~Qian and R.~J. et~al., ``Efficient distance metric learning by adaptive
  sampling and mini-batch stochastic gradient descent {(SGD)},'' {\em Mach.
  Learn.}, vol.~99, no.~3, pp.~353--372, 2015.

\bibitem{DBLP:journals/corr/GunayAC14}
O.~G{\"{u}}nay and C.~E.~A. et~al., ``Cosine similarity measure according to a
  convex cost function,'' {\em Arxiv}, p.~abs/1410.6093, 2014.

\bibitem{DBLP:journals/bstj/Shannon48}
C.~E. Shannon, ``A mathematical theory of communication,'' {\em Bell Syst.
  Tech. J.}, vol.~27, no.~3, pp.~379--423, 1948.

\bibitem{DBLP:conf/icml/ZhengMWCYML17}
S.~Zheng and Q.~M. et~al., ``Asynchronous stochastic gradient descent with
  delay compensation,'' in {\em International Conference on Machine Learning,
  ICML}, p.~70, 2017.

\bibitem{DBLP:journals/corr/abs-1708-07747}
H.~Xiao and K.~R. et~al., ``Fashion-mnist: a novel image dataset for
  benchmarking machine learning algorithms,'' {\em Arxiv}, p.~abs/1708.07747,
  2017.

\bibitem{DBLP:journals/corr/abs-1811-07270}
T.~Ho{-}Phuoc, ``{CIFAR10} to compare visual recognition performance between
  deep neural networks and humans,'' {\em Arxiv}, p.~abs/1811.07270, 2018.

\bibitem{JMLR:v24:22-0440}
D.~Zeng and S.~L. et~al., ``Fedlab: A flexible federated learning framework,''
  {\em Journal of Machine Learning Research, JMLR}, vol.~24, no.~100, pp.~1--7,
  2023.

\bibitem{DBLP:conf/nips/KrizhevskySH12}
A.~Krizhevsky and I.~S. et~al., ``Imagenet classification with deep
  convolutional neural networks,'' in {\em Neural Information Processing
  Systems, NeurIPS}, 2012.

\end{thebibliography}

\clearpage
\appendix
\section{Implementation Details}
    \textbf{Baselines.} We compare FedCME with several advanced methods, including FedAvg \cite{DBLP:conf/aistats/McMahanMRHA17}, FedProx \cite{DBLP:conf/mlsys/LiSZSTS20}, MOON 
    \cite{DBLP:conf/cvpr/LiHS21} and FedRS \cite{DBLP:conf/kdd/LiZ21}.
    FedProx uses a proximal term to reduce the gradient variance.
    MOON  adds a model-contrastive loss to control the training direction of the model.
    FedRS adds weight parameters to softmax layer to limit it to update inaccurate directions.

    \textbf{Dataset.} FMNIST \cite{DBLP:journals/corr/abs-1708-07747} and CIFAR10 \cite{DBLP:journals/corr/abs-1811-07270}  with
    heterogeneous dataset partition are used to test the efficacy
    of FedCME, which  are widely adopted in FL research. Same as previous works \cite{DBLP:conf/iclr/LiHYWZ20,DBLP:conf/nips/LuoCHZLF21, JMLR:v24:22-0440}, we use Dirichlet distribution~$dir(\alpha)$ on label radios to simulate the heterogeneous data distribution among clients, where a smaller $\alpha$ indicates higher data heterogeneity. During the implementation, we set $\alpha$ = 0.1 and $\alpha$ = 0.5.

    \textbf{Hyperparameters Settings.} In order to imitate the environment of real FL environment where there are many clients with only a small number of samples each, we set up two levels of client quantity $|\mathcal{K}|=50$ and $|\mathcal{K}|=80$. Based on the above, we use four client selection options. For $|\mathcal{K}|=50$, $|\mathcal{M}|$=10 or 20, and mini-batch size $|\mathcal{B}|$ is 32. For  $|\mathcal{K}|=80$,  $|\mathcal{M}|$=20 or 30, and $|\mathcal{B}|$ is 64. And for all methods, learning rate $\eta$ is 0.01 and local train epoch $E$ is 6. Especially, the regularization term factor in FedProx we set 0.1 and 0.01. And the factor in FedRS we set 0.1 and 0.5 as recommended. For our method FedCME, we set $\mu$=0.1 or 0.01. After local training, we take $20\%$ of the local data as the evaluation dataset.

    \textbf{Network Architecture.} For FMNIST, we use a simple convolutional neural network(CNN) model which is composed of two convolutional layers and two linear layers. For CIFAR10, we employ AlexNet \cite{DBLP:conf/nips/KrizhevskySH12} as the basic backbone. We divide the final linear layers of the model into the classifier, and the network model layers before it is divided into feature extractor.

    \textbf{Metrics.} Our objective is to minimize the empirical loss during the training process, and to train a global model with better performance. Therefore, the efficiency of the proposed algorithm is quantified as the test accuracy under different degrees of data heterogeneity and the client selection strategies.

\end{document}